\documentclass[conference]{IEEEtran4PSCC}

\IEEEoverridecommandlockouts
\usepackage{cite}
\usepackage{amsmath,amssymb,amsfonts}
\usepackage{algorithmic}
\usepackage{graphicx}
\usepackage{textcomp}
\usepackage[caption=false,font=footnotesize]{subfig}
\usepackage{color}
\usepackage{xcolor}
\usepackage{cite}
\usepackage{amsmath}
\usepackage{amssymb}
\usepackage{amsthm}
\usepackage{graphicx}

\usepackage{epstopdf}
\usepackage{comment}
\usepackage{multirow}
\usepackage{paralist}
\usepackage{lineno}
\usepackage{mdwlist}
\usepackage{eurosym}\DeclareGraphicsExtensions{.pdf,.png,.jpg}
\usepackage{breqn}
\usepackage{makecell}
\usepackage{soul} 
\graphicspath{{./pic/}}

\usepackage{subfig}
\usepackage{bm}
\usepackage[normalem]{ulem}
\usepackage{xcolor}
\def\BibTeX{{\rm B\kern-.05em{\sc i\kern-.025em b}\kern-.08em
    T\kern-.1667em\lower.7ex\hbox{E}\kern-.125emX}}

\begin{document}

\title{Wind Power Scenario Generation Using Graph Convolutional Generative Adversarial Network}

\author{\IEEEauthorblockN{Young-ho Cho, Shaohui Liu, and Hao Zhu}
\IEEEauthorblockA{Chandra Department of Electrical and Computer Engineering \\
The University of Texas at Austin\\
Austin, TX, USA \\
\{jacobcho, shaohui.liu, haozhu\}@utexas.edu}
\and
\IEEEauthorblockN{Duehee Lee}
\IEEEauthorblockA{Department of Electrical Engineering \\
Konkuk University\\
Seoul, South Korea \\
hello.keanu@utexas.edu}
}

\maketitle

\begin{abstract}
    Generating wind power scenarios is very important for studying the impacts of multiple wind farms that are interconnected to the grid.
    We develop a graph convolutional generative adversarial network (GCGAN) approach by leveraging 
	GAN's capability in generating  large number of realistic scenarios without using statistical modeling. Unlike existing GAN-based wind power data generation approaches, we design GAN's hidden layers to match the underlying spatial and temporal characteristics.
	We advocate the use of graph filters to embed the spatial correlation among multiple wind farms, and a one-dimensional (1D) convolutional layer to represent the temporal feature filters.  The proposed graph and feature filter design significantly reduce the GAN model complexity, leading to improvements in training efficiency and computation complexity. 
	Numerical results using real wind power data from Australia demonstrate that the scenarios generated by the proposed GCGAN exhibit more realistic spatial and temporal statistics than other GAN-based outputs.
\end{abstract}


\thanksto{\protect\rule{0pt}{0mm}
This work has been supported by NSF Grants 1802319 and 2130706.}

\section{Introduction}

The growth of wind generation and other carbon-free sources increasingly challenges power system operations with high uncertainty and variability \cite{bessa2014handling}. Due to the dependence on weather and atmospheric conditions, accurate predictions of wind speed and direction are difficult to achieve \cite{bokde2019review}, critically affecting the integration of wind power generation. It is imperative to improve the modeling and understanding of wind farm outputs at specific geo-locations for attaining reliable  and economic power system operations and planning.


To generate synthetic wind power scenarios, a generative adversarial network (GAN) approach has been recently advocated  that can eliminate the need of statistical assumptions.
Traditional time-domain statistical models, such as autoregressive moving average (ARMA) \cite{morales2010methodology} and generalized dynamic factor  \cite{lee2016load}, require numerical and structural assumptions, and thus they may fail to capture all possible underlying factors that contribute to wind power variability.
Meanwhile, the GAN approach generates fake datasets by directly imitating the original data and their marginal distributions via the interactive improvements between a generator and discriminator~\cite{goodfellow2014generative}.
The work \cite{chen2018model} first explored the use of GAN for generating wind and solar power scenarios, whereas the vanilla model structure therein was unable to fully capture the intrinsic characteristics of renewable sources.  The sequence GAN was developed by \cite{liang2019sequence} using a special temporal filter design.
Convolutional neural network (CNN) was used in~\cite{wang2022approach}, with the goal of attaining the correct spatial correlation among multiple wind farms.
While the spatial convolution based on a rigid 2-dimensional (2D) grid of CNN is useful to  match the local correlation with neighboring nodes,  it could unfortunately fail to represent the global correlation among any pair of wind farms.

In this paper, we suggest an enhancement of the GAN-based wind power generation approach by using graph convolutional network (GCN) to produce the correct spatial relations among multiple wind farms. GCN uses graph filters to mix the input features among a network of nodes, and this graph convolution can efficiently incorporate the underlying node connectivity and graph embedding~\cite{kipf2016semi,garg2020generalization}. By viewing the power system as a graph, GCN has successfully incorporated the grid topology in solving the problems of predicting the electricity market prices~\cite{liu2021graph}, locating line faults~\cite{chen2020gcnfault}, and learning the optimal power flow solutions~\cite{liu2022topology}.
To develop the proposed GCGAN approach, we model the hidden layers of GAN's generator and discriminator to consist of both  graph filters in spatial dimension and feature filters in temporal dimension. 
To match with the spatial correlation of given wind farms, the graph filter weights are obtained from the correlation coefficients through an exponential relation.  
This filter design can guide the GAN training process to recognize both strong and weak correlations among wind farms.
Due to the high-dimensionality of the scenario generation window, the temporal feature  filter of GCGAN is simplified by using  a one-dimensional (1D) convolutional layer. This design can greatly reduce the number of trainable weight parameters and thus reduce the computation needs of training GCGAN.
In order to implement the proposed GCGAN approach, we use real wind power data from multiple wind farms in Australia. We perform numerical comparisons to validate its effectiveness in terms of producing consistent spatial correlation and temporal variability.

Our contributions are as follows:
\begin{enumerate}
	\item We build a wind power scenario generation scheme  that accounts for the spatial and temporal characteristics of wind farm outputs;
	\item Our proposed graph and temporal feature filters are greatly simplified to attain good computational efficiency.
\end{enumerate}

The rest of the paper is organized as follows.  Section~\ref{sec:gan} provides an overview of the generative adversarial network (GAN) based wind power scenario generation framework.
In Section~\ref{sec:gcgan}, we develop the graph convolutional GAN (GCGAN) model and discuss the design of its graph and feature filters that are consistent with the spatial and temporal characteristics of wind data.
Section~\ref{sec:sim} presents the numerical comparisons and validations for the proposed scheme, along with some concluding remarks.


\section{Wind Scenario Generation Using GAN}
\label{sec:gan}

We provide an overview of the generative adversarial network (GAN) model used for developing the proposed scenario generation framework.
Several factors need to be considered when generating wind power scenarios. 
First, the power output data of an individual wind farm exhibit unique temporal characteristics due to the wind speed variability and other plant operating conditions.
Second, the spatial correlation among multiple wind farms is important to maintain especially for considering their joint impact on the interconnected power grid.
It is important to account for these complicated characteristics in the design of wind power scenario generation methods.

The GAN model~\cite{goodfellow2014generative} is powerful in generating synthetic data with similar characteristics as the given real dataset, without the need of explicitly modeling the underlying data statistics. Although it was mainly developed for generating fake images or videos, this method could also be used for generating synthetic time series in power systems. 
The GAN architecture consists of a generator $G$ and a discriminator $D$, both of which are essentially neural networks (NNs). The generator $G$ outputs the synthetic data samples, while the discriminator $D$ determines whether they are realistic or not using the given real data samples. This resultant binary classification output will be further incorporated by the generator $G$ to improve the data generation performance. Note that the discriminator $D$  performs a classification by using the input real data samples as a supervised learning task. In contrast, the generator $G$ does not have direct access to the real data input.


The gist of GAN prediction is to gradually improve the data generation performance through the interaction between $G$ and $D$. While $D$ tries to maximize its own performance in classifying fake data from real data, $G$ instead intends to reduce  this classification performance by producing more realistic data. Therefore, the overall problem can be cast as a min-max problem that iteratively updates the NN parameters of $G$ and $D$ until an equilibrium is attained.
Suppose we want to generate the wind power scenarios for $N$ wind farms and  $T$ time steps. The generator $G$ produces a fake data sample $\hat {\bm X} \in \mathbb R^{N \times T}$ using a random noise matrix $\bm Z \in \mathbb R^{N \times K}$, as given by
\begin{equation}
	\hat {\bm X} = G(\bm Z).
\end{equation}
The noise matrix $\bm Z$ is sampled from a known probability distribution $P_{\bm Z}$, such as Gaussian distribution or Laplace distribution. The dimension $K$ of input noise is chosen to be much smaller than $T$. The goal is for the generated sample $\hat {\bm X}$ to approach the probability distribution of the real data samples, as denoted by $P_{\bm X}$. Accordingly, the discriminator classifies on the authenticity of $\hat {\bm X}$. Its output $D(\hat {\bm X})=1$ if the discriminator decides it is real data; and 0 otherwise. Based on the cross-entropy loss, the objective for $G$ is to minimize
\begin{equation}
	\label{gen_GAN}
	\mathcal{L}_G = \mathbb{E}_{\bm Z \sim P_{\bm Z}} ~\log [1-D(G(\bm Z ))]
\end{equation}
where $\mathbb{E}$ denotes the expected operator. 
Clearly, if the discriminator $D$ is more likely to misclassify the generated sample $\hat {\bm X}$ as real data, then the term $\mathcal{L}_G$ becomes smaller. 
Meanwhile, for the classification task of $D$, the objective is to maximize 
\begin{align}
	\mathcal{L}_D &= \mathbb{E}_{\bm X \sim P_{\bm X} } \log [D(\bm X)] \nonumber\\ &\qquad +\mathbb{E}_{\bm Z \sim P_{\bm Z} } \log [1-D(G(\bm Z))]. \label{dis_GAN}
\end{align}
By combining these two individual objectives, the generator $G$ and discriminator $D$ jointly solves the following min-max problem, as
\begin{align}
	\min\limits_{G} \max\limits_{D} ~~~& \mathbb{E}_{\bm X \sim P_{\bm X} } \log [D(\bm X)] \nonumber\\ & +\mathbb{E}_{\bm Z \sim P_{\bm Z}} \log [1-D(G(\bm Z))]. \label{obj_GAN}
\end{align}
By using gradient updates, this min-max problem can be solved to obtain the NN parameters for $G$ and $D$. 

The design of GAN's hidden layer structures is important for attaining consistent characteristics with the real data samples. It is possible to use 2D CNN that lead to correlated outputs from multiple locations~\cite{wang2022approach}. Specifically, by assigning the wind farms to the locations on a 2D grid, the convolutional filters aggregate the input data from adjacent locations and can create correlated outputs within the local regions.  
Nonetheless, the local patterns produced by CNN-based scenario generation approaches could fail to capture the globally correlated relations among all wind farm locations, due to the limitations on the 2D filter size. For example, the level of correlation between two wind farms is difficult to maintain if they are not adjacent to each on the 2D grid. 
Thus, the ensuing section will develop a  graph convolutional layer based approach to improve the consistency spatial correlation in the wind power scenarios generated by GAN.

\begin{figure}[t!]
	\begin{center}
		\includegraphics[scale=0.35]{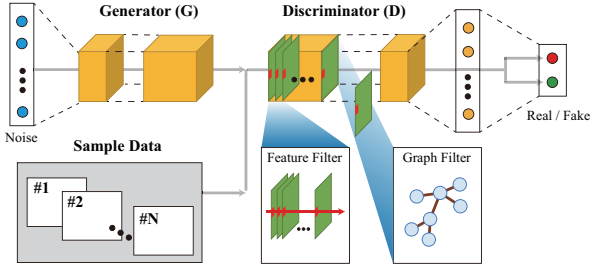}
		\caption{The structure of the proposed graph convolutional generative adversarial network (GCGAN).} \label{GAN}
	\end{center}
\end{figure}

\section{Graph Convolutional GAN (GCGAN)}
\label{sec:gcgan}

The key of GCGAN lies in using a graph convolutional network (GCN) approach to model the layers in the GAN's generator and discriminator, as illustrated in Fig.~\ref{GAN}.
The GCN is a special type of NNs that can effectively predict nodal signals that have a graph-based dependence. Specifically, each layer of GCN uses a graph filter that embeds the underlying dependence among the set of nodes that can aggregate the input signals from other connected nodes. This way, the output data of GCN would exhibit strong correlation among the nodes that are connected to each other.   
We will design the graph filter $\bm A \in \mathbb R^{N \times N}$ later on  to match the actual spatial correlation among the $N$ wind farms.

Let $L$ denote the total number of hidden layers in the generator,  and $\bm X^{(\ell)}\in \mathbb R^{N\times K_{\ell}}$ represent the input data matrix with $K_\ell$ features per layer $\ell\in\{1,\dots, L\}$. The input to the first layer is essentially the random noise matrix; i.e.,  $\bm X^{(1)} = \bm Z$ and $K_1 = K$. Note that we set $K_{\ell+1} > K_\ell$ in order to gradually increase the temporal dimension. For each layer $\ell$, using the matrix 
$\bm W^{(\ell)} \in \mathbb R^{K_{\ell} \times K_{\ell+1}}$ as the trainable feature filter parameters, we can represent it as
\begin{align}
	\bm X^{(\ell+1)} = \sigma(\bm A \bm X^{(\ell)} \bm W^{(\ell)}),\, \ell=1, \dots, L \label{GCN_layer}
\end{align}
where $\sigma(\cdot)$ denotes the nonlinear activation function.
We can pick \texttt{ReLU} as the activation function of most layers,  while the last layer $L$ uses the hyperbolic tangent function  \texttt{tanh} to produce output values within $[-1,1]$ as wind power data. Clearly, the graph filter $\bm A$ in \eqref{GCN_layer} plays the important role of aggregating the nodal features such that the rows of the final layer's output  $\bm X^{(L+1)} = \hat{\bm X}$ can exhibit the needed correlation pattern. 

As for the layers of discriminator, they are constructed very similarly to \eqref{GCN_layer}, with the first layer using $\hat{\bm X}$ as the input. A difference is that the number features $K_{\ell}$ decreases with $\ell$, as this is a binary classification task and the last layer outputs the binary decision. As for its activation functions,  most of the layers use \texttt{Leaky ReLU},  and for the last layer the \texttt{sigmoid} function is selected to produce values within $[0,1]$ for binary classification. 

\subsection{Graph Filter for Spatial Dimension}

The choice of GCN's graph filter $\bm A$ can critically affect the spatial correlation of the generated data $\hat{\bm X}$.  
Note that some GCN-based approaches, such as \cite{liu2022topology}, have included the graph filters as trainable parameters for attaining high prediction accuracy, but this setting may increase the computation complexity and possibly lead to overfitting. As the correlation matrix $\bm C$ can be estimated accurately from the given real wind data, we will use $\bm C$ to fix the graph filter weights throughout all layers. For example, earlier work on using GCN for forecasting wind and solar generation~\cite{liao2022short} has directly used the absolute value of correlation  coefficients by setting $\bm A = |\bm C|$, and similarly in other GCN work \cite{hou2022gcns}. 
Albeit simple, this choice could be subpar for attaining the actual spatial correlation. This is because the  weak correlation level (0$\sim$0.3) and  strong correlation one (0.7$\sim$1.0) cannot be easily differentiated based on the absolute value itself.

To this end, we propose to use an exponential transformation on the correlation coefficients, by setting the $(i,j)$-th entry of the graph filter as
\begin{align}
	\bm A_{i,j} = 
	\left(\frac{e^{|\bm C_{i,j}|} -1}{e-1}\right)
\end{align}
where the fractional term is normalized to be within $[0,1]$.
As shown in \cite{de2020edge}, this exponential transformation could enhance the significance of filter weights between strongly correlated locations.
Our proposed graph filter design turns out effective in matching with the pairwise correlation for the wind farms.

\subsection{Feature Filter for Temporal Dimension}

The large size of temporal feature filters $\{\bm W^{(\ell)}\}$ makes it inefficient to train all these parameters during the GAN training process. This is because wind power scenarios are typically generated for a month or even a year, and thus the feature dimension $K_\ell$ would eventually be very large. Thus, it is necessary to simplify the temporal filter design in order to prevent data overfitting. 

To this end, we design the feature filter by utilizing the underlying temporal correlation in the wind power time series. By using a 1D convolutional filter that sequentially strides the input features, we can significantly reduce the number of parameters while maintaining the temporal correlation. To implement this idea, consider the matrix  multiplication term $\tilde{\bm X}^{(\ell)} = \bm X^{(\ell)}\bm W^{(\ell)}$ in \eqref{GCN_layer} as the output of feature filtering. Instead of having a full matrix $W^{(\ell)}$, we  use a 1D convolution filter of length $(2M_{\ell}+1)$ with trainable weight coefficients $\{W_{-M}^{(\ell)}, \ldots, W_0^{(\ell)}, \ldots, W_M^{(\ell)}\}$. This way, the $j$-th column of matrix $\tilde{\bm X}^{(\ell)} $ can be formed by 
\begin{align}
    \bm {\tilde{X}}^{(\ell)}_{:,j} &= \sum_{m=-M_{\ell}}^{M_{\ell}} \bm X^{(\ell)}_{:,j-m} W^{(\ell)}_m,  
\end{align}
and this filter output is used to construct each GCGAN layer as $\bm X^{(\ell+1)} = \sigma(\bm A \bm {\tilde{X}}^{(\ell)})$.
This way, the full filter matrix is reduced to a small 1D filter of fixed length. The latter is still able to match the temporal correlation patterns of the wind power scenarios while reducing the complexity burden of training process. This completes the design of our proposed GCGAN-based wind power scenario generation scheme.


\begin{table}[t!]
	\caption{Comparison of the training time per epoch and the total training time for convergence.}
	\begin{center}
		\begin{tabular}{c|c|c|c}
			\Xhline{3\arrayrulewidth}
			& DCGAN & GCGAN\_full & GCGAN\\
			\hline
			\makecell[c]{Training Time\\per epoch} & 6.21 s & 4.17 s & 1.31 s \\
			\hline
			\makecell[c]{Convergence Time} & 594.7 s & 414.2 s & 68.3 s \\
			\Xhline{3\arrayrulewidth}
		\end{tabular} 
		\label{Training_time}
	\end{center}
\end{table}

\section{Numerical Results}
\label{sec:sim}
We have implemented the proposed GCGAN scheme  using historical wind power data from $N=20$ wind farms that have registered with the Australian Energy Market Operator (AEMO). The wind power outputs of each wind farm were collected at 5-minute intervals from 2012 to 2017. The numerical implementation has been performed in PyTorch on a regular laptop with Intel\textsuperscript{\textregistered} CPU @ 2.70 GHz, 32 GB RAM, and NVIDIA\textsuperscript{\textregistered} RTX 3070 Ti GPU @ 8GB VRAM.

To demonstrate the efficiency of the proposed 1D feature filter, we also consider the GCGAN model with the full matrices $\bm W$'s as feature filters, termed as GCGAN\_full. In addition, a CNN-based model using  deep convolutional GAN (DCGAN) has been implemented by arranging the 20 wind farms into a 2D grid based on the latitude and longitude and trained as a 3D CNN model. For all models, the scenario is generated for a 10-day duration of $T=2880$ time steps. The GAN input noise dimension is $K=5$, and number of hidden layers for either the generator or discriminator is $L=4$.
The generator of the GCGAN models has $K_\ell =\{5,180,720,2880\}$ features at each layer $\ell$, and the discriminator has $K_\ell=\{2880,720,180,5\}$ features. The size parameter for GCGAN's 1D temporal feature filters is set to $M_{\ell}=\{12,72,144,144\}$,  corresponding to the dynamics at 1-hour, 6-hour, or 12-hour intervals.


We first compare the computational time for the proposed GCGAN with the other two GAN-based models.
Table~\ref{Training_time} lists the training time per epoch and the total training time for convergence. Clearly, the proposed GCGAN based on temporal filter is significantly faster than GCGAN\_full using fully-connected feature filters.
Moreover, both of them are faster than the CNN-based DCGAN scheme.
The proposed GCGAN model also takes less time for convergence compared to other GAN-based models. 
It is true that there is a possibility of underfitting due to a small number of parameters in GCGAN. But we have ensured to maintain the spatial and temporal characteristics by carefully designing graph and feature filters while attaining an accelerated convergence time. We will further demonstrate the improvement of GCGAN in both spatial and temporal domains.

\begin{figure}[t!]
	\centering
    \subfloat[Actual data]{\includegraphics[scale=0.17]{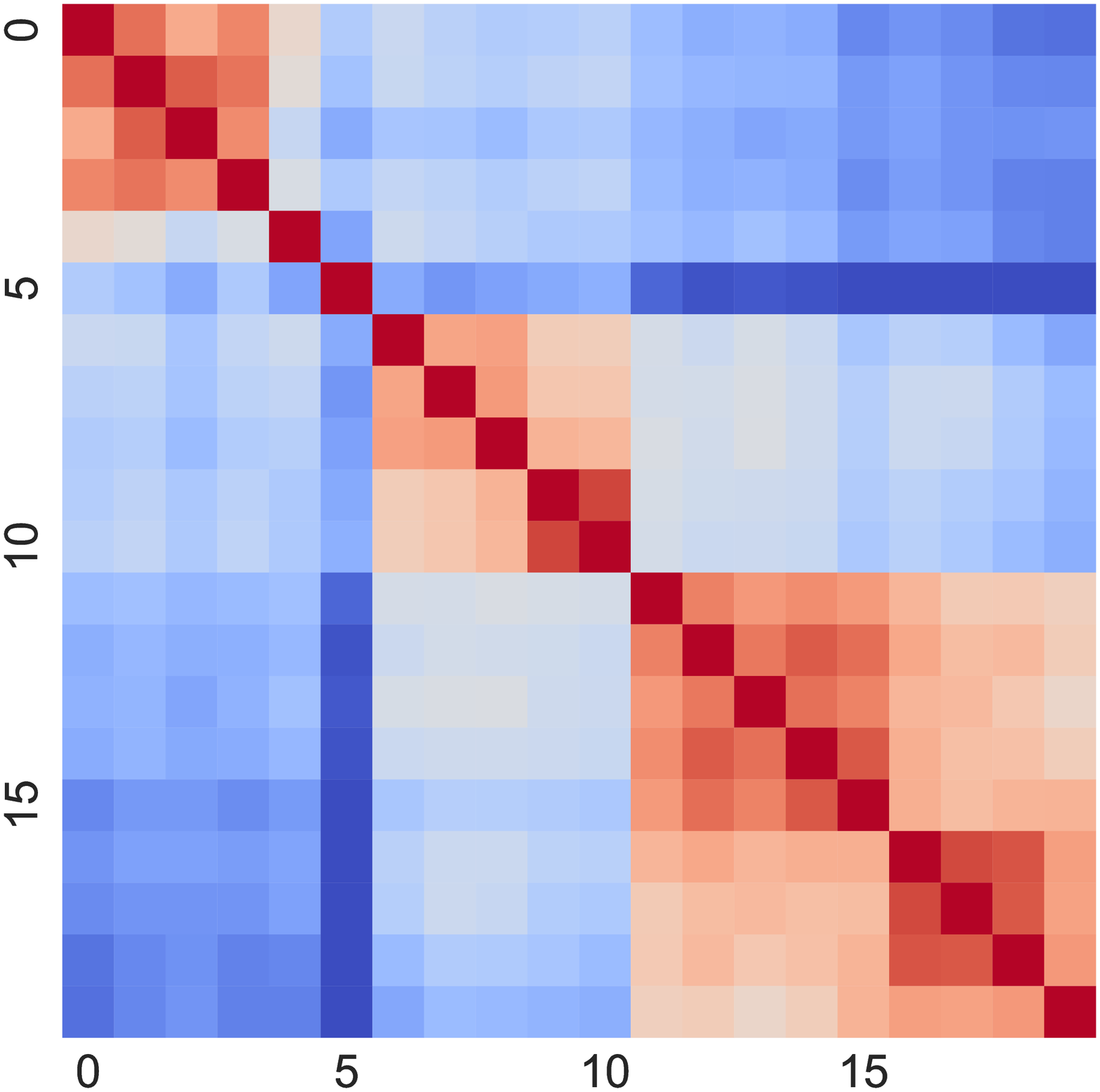}\label{Cov_AU_act}}
	\quad
	\subfloat[DCGAN]{\includegraphics[scale=0.17]{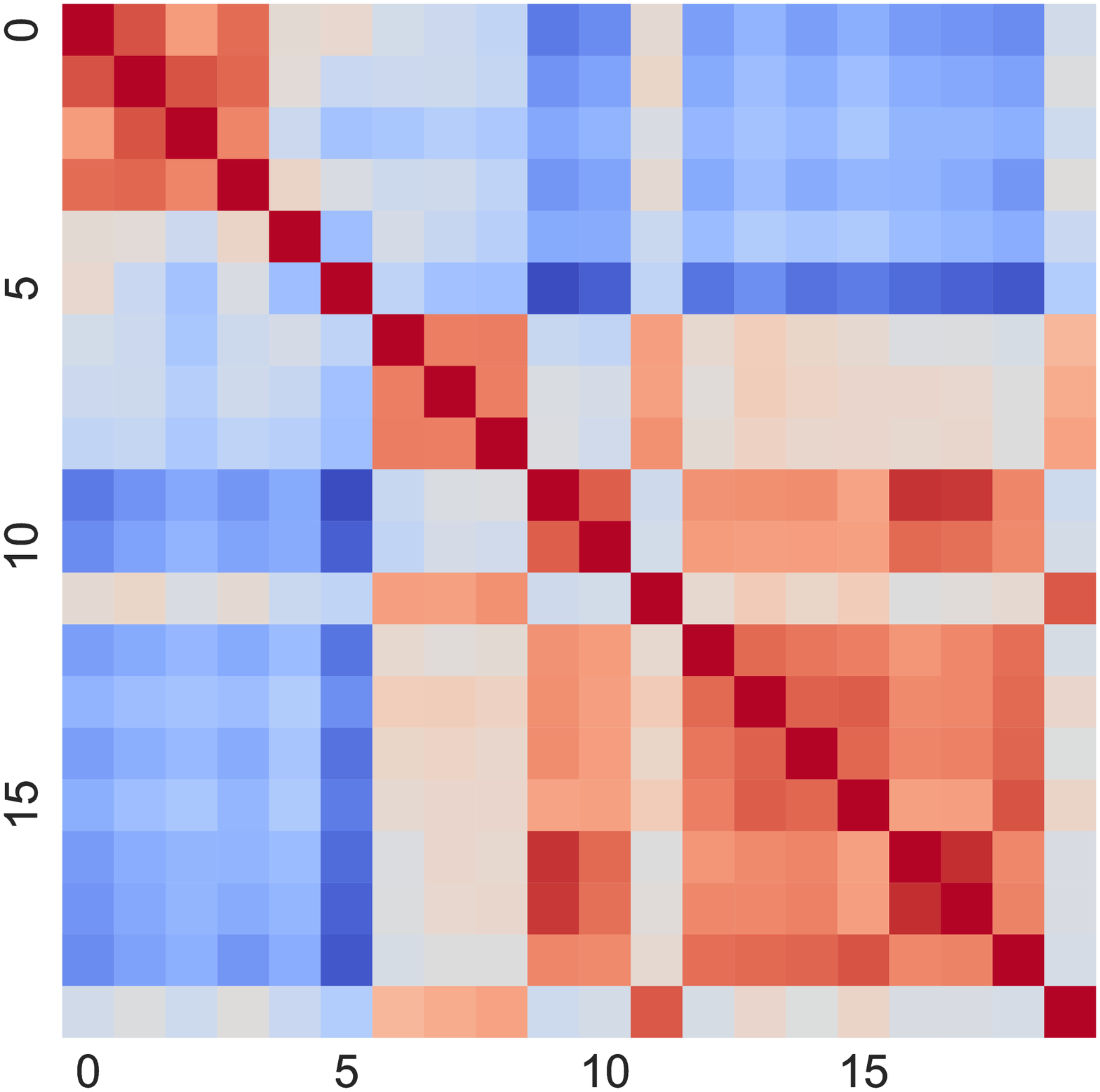}\label{Cov_DCGAN}}
	\quad
	\subfloat[GCGAN\_full]{\includegraphics[scale=0.17]{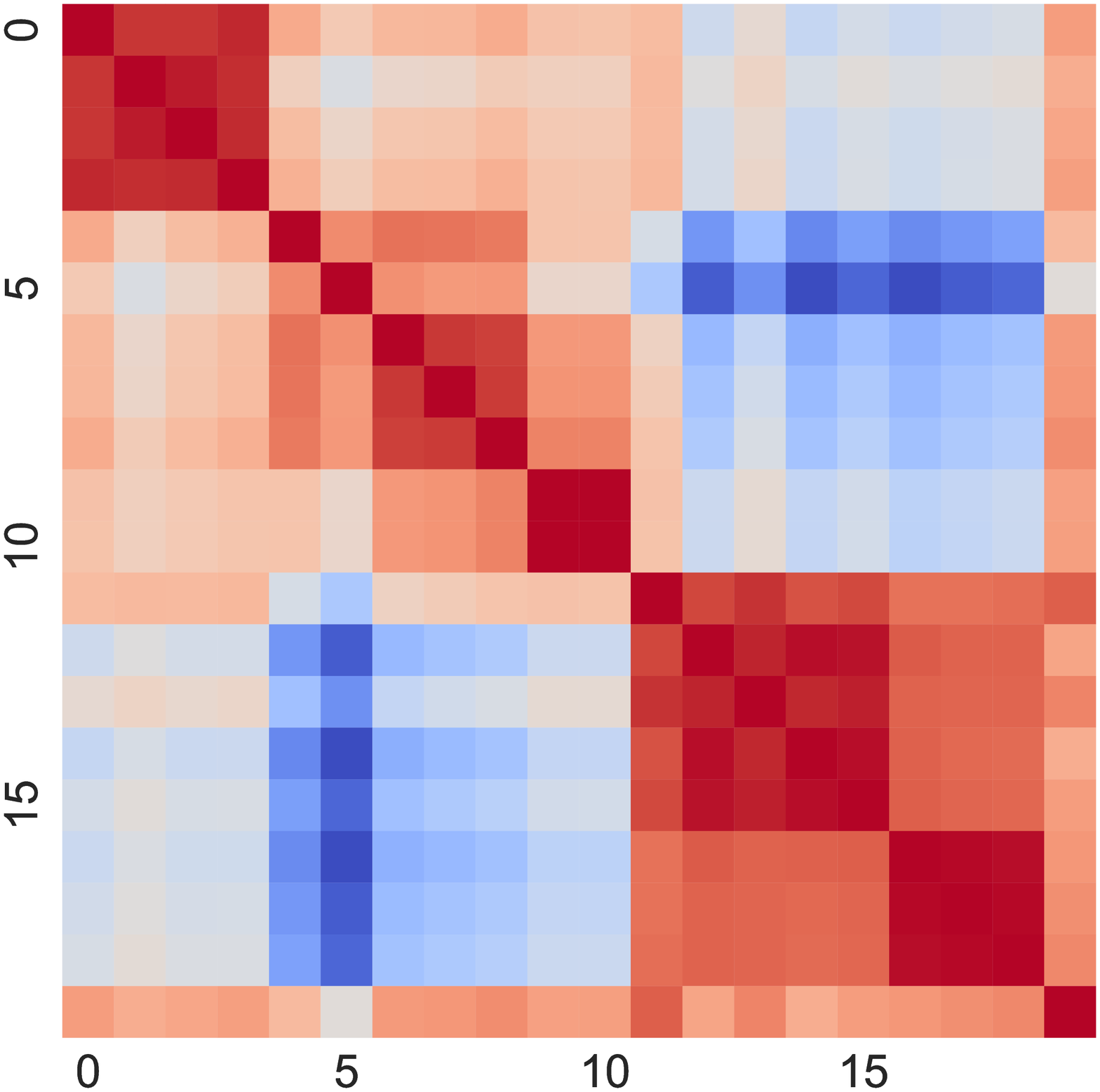}\label{Cov_GCGAN}}
	\quad
	\subfloat[GCGAN]{\includegraphics[scale=0.17]{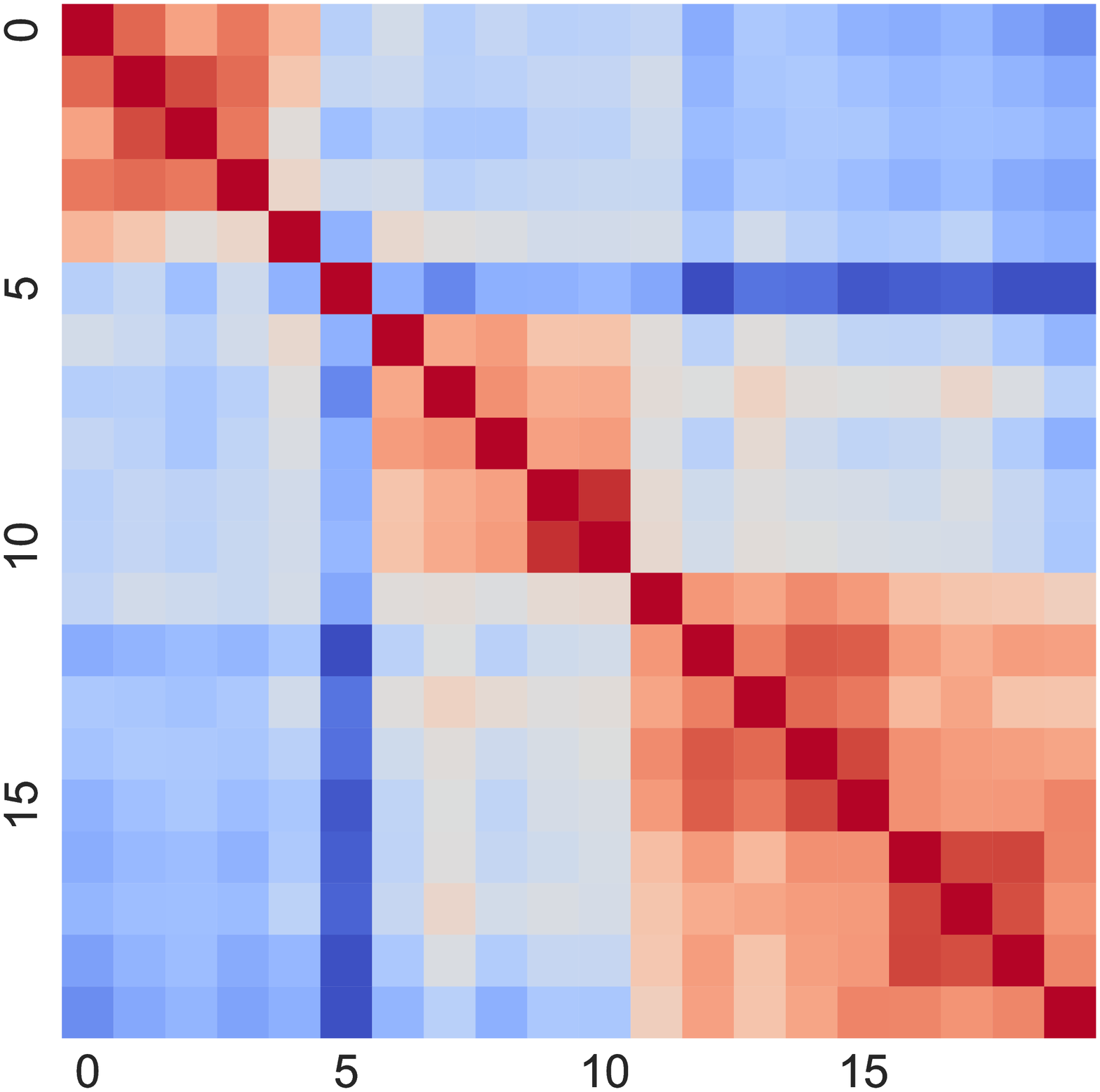}\label{Cov_GCGAN2}}
    \quad
    \subfloat{\includegraphics[scale=0.25]{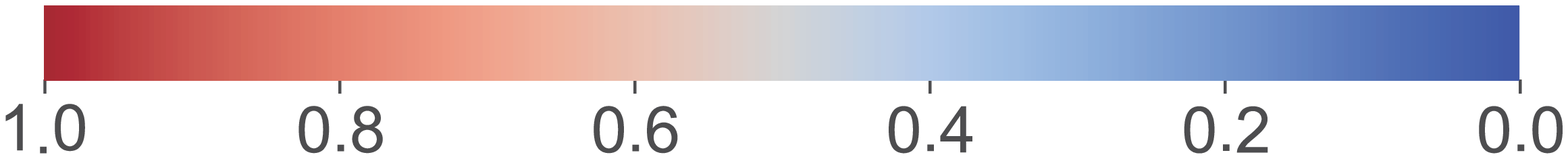}\label{cbar}}
	\caption[]{\small Comparison of the correlation matrices of the actual and generated scenarios by different GAN models.}\label{Cov}
\end{figure}

\begin{figure}[t!]
	\centering
	\subfloat{\includegraphics[scale=0.3]{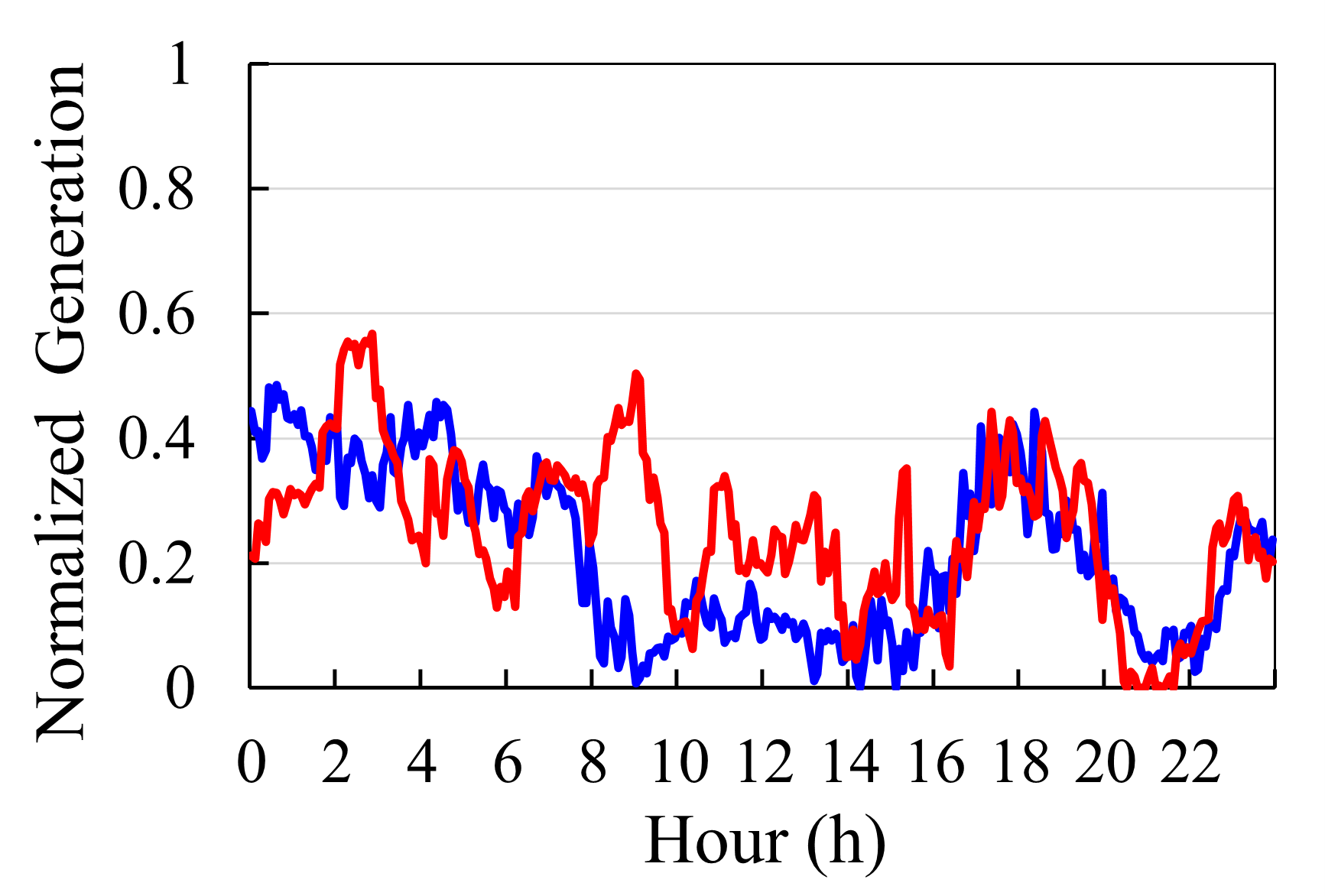}\label{DCGAN_time}}
	\quad
	\subfloat{\includegraphics[scale=0.3]{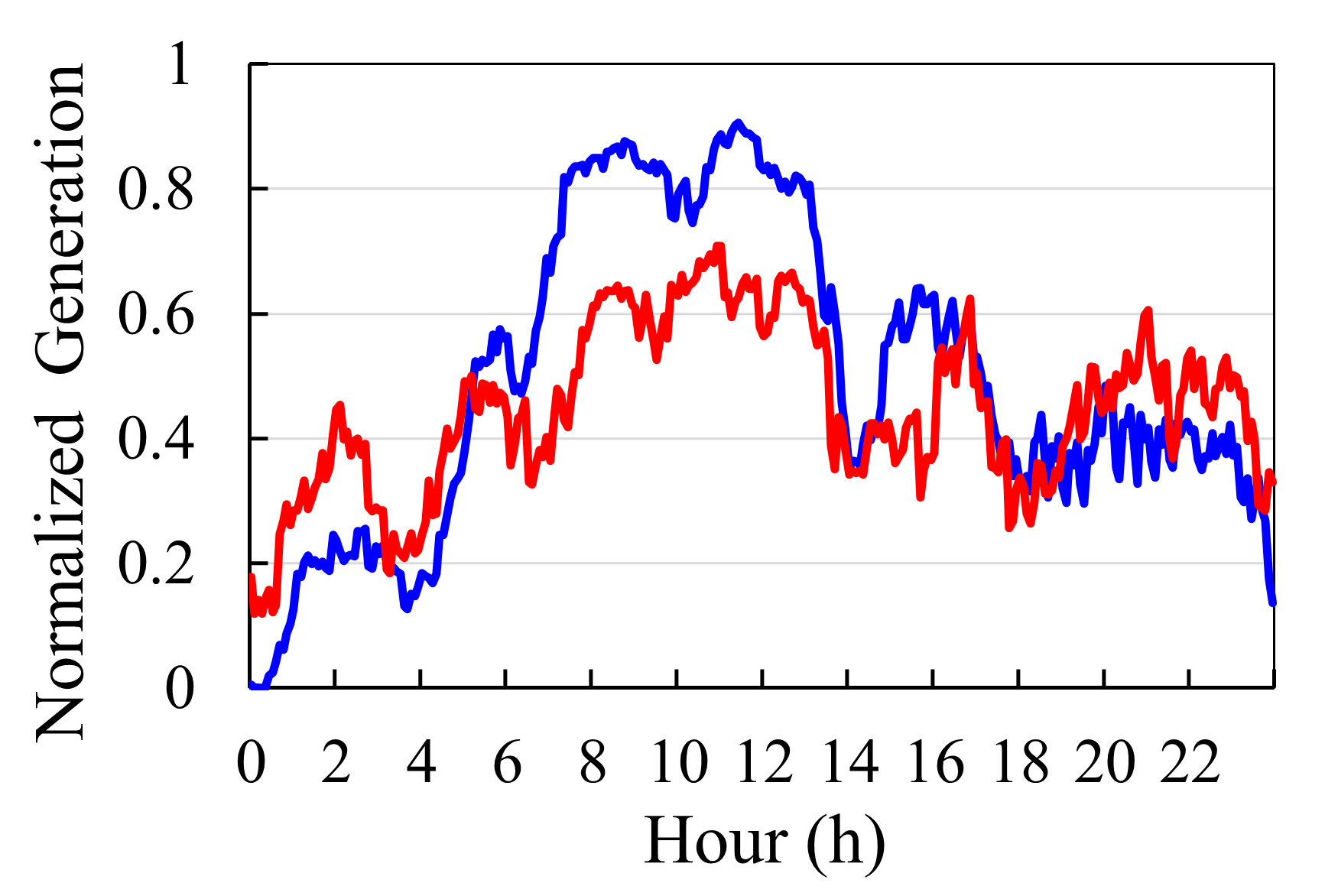}\label{GCGAN_time}}
	\caption[]{\small Comparison of sample wind power scenarios generated by DCGAN (left) and GCGAN (right) for two  selected wind farms that are geographically close and highly correlated.}\label{timeseries}
\end{figure}

\subsection{Spatial Correlation} 

We compare the spatial correlation of the data generated by each  GAN model with that of the actual historical data.
Fig.~\ref{Cov} plots the correlation matrices with the entry values lying within $[0,1]$, as all correlation coefficients turn out to be non-negative. For the actual data, the nearly-red diagonal blocks show several clusters of highly-correlated wind farms. The wind farms within each cluster are geographically much closer to each other than to those in other clusters.

For the scenario generated by each GAN-based model, the proposed GCGAN shows the highest similarity in recovering the actual correlation. In particular, GCGAN  can maintain the correlation level for wind farms that are either  strongly or weakly correlated. As for the DCGAN one, the correlation level has significantly increased for some weakly correlated wind farms or those not in the same cluster. This is likely due to the 2D convolutional layer design which has created some unnecessary correlation between certain pairs of wind farms. The same observation holds for the GCGAN\_full one, which shows the correlation level could be excessively high. We believe this is because of its learning inefficiency as a result of using the high-dimensional feature filters. Different from these two models, the proposed GCGAN achieves a good balance to maintain both strong and weak correlation levels among the wind farms. As an example, Fig.~\ref{timeseries} compares sample daily outputs generated by DCGAN and GCGAN for two selected wind farms whose geographical proximity has led to a high level of correlation at {0.925}. While the DCGAN outputs show an opposite trend at certain hours, the GCGAN ones maintain high  similarity throughout the day.
While there could be some space for improvement, the proposed exponential rule-based graph filter design turns out useful to produce the correct spatial correlation, which the simple temporal filter has made the optimization of GAN parameters very effective.

\subsection{Temporal Variability} 

Furthermore, the  temporal variability of the generated scenarios by each GAN model is compared with that of the actual data. Basically, this characteristic represents the change rate of a wind farm output, as defined by the wind power difference between the beginning and end of a fixed time interval. We have picked the 15-min, 30-min, and 60-min variability values for comparison, with the statistics listed in Table~\ref{stat_table} for both the proposed GCGAN generated data and the actual data. 
The variability values of wind farm output typically follow a Laplace distribution, and thus we have compared the peak value and variance of the variability distribution. 
While both statistics for the 15-min variability are slightly higher than the actual values, there is a pretty good match between the generated scenario and actual data.
Therefore, the proposed scenario generation method creates scenarios with similar variability to the actual data.

\begin{table}[t!]
	\caption{Comparison of the statistical properties of the actual data and GCGAN generated scenarios.}
	\begin{center}
		\begin{tabular}{c|c|c|c}
			\Xhline{3\arrayrulewidth}
			Types & \makecell[c]{Time\\Interval} & \makecell[c]{Actual\\Data} & \makecell[c]{Generated\\Scenarios} \\
			\hline
			\multirow{3}{*}{\makecell[c]{Variability\\Peak}} & 15-min & 0.1546 & 0.2051  \\ 
			 & 30-min & 0.1046 & 0.0835 \\
			 & 60-min & 0.0764 & 0.0720 \\
			\hline
			\multirow{3}{*}{\makecell[c]{Variability\\Variance}} & 15-min & 3.019e-04 & 5.686e-04 \\
			& 30-min & 1.577e-04 & 1.544e-04 \\
			& 60-min & 1.055e-04 & 1.302e-04 \\
			\Xhline{3\arrayrulewidth}
			Plant CF & & 35.2\% & 35.3\% \\
			\hline
			\makecell[c]{Weibull\\Parameter} & & (0.3281, 2.92) & (0.3614, 2.98) \\
			\Xhline{3\arrayrulewidth}
		\end{tabular} \label{stat_table}
	\end{center}
\end{table}

In addition to the variability statistics, Table~\ref{stat_table} further compares the plant capacity factor (CF) and distribution parameters, both related to the wind farm's annual production pattern. They are very important for using the generated wind scenarios in power system planning studies.
Given the total annual power generation of $P$, the plant CF is defined by:
\begin{align}
\alpha_C = P/(C *24*365)
\end{align}
where $C$ denotes its installed capacity. In general, the CF ratio is between 25\% and 40\%~\cite{ackermann2012wind}, and 
 shows that the generated data has a CF of 35.3\% which very matches the actual value at 35.2\%. 
 Furthermore, as the wind power output typically follows a Weibull distribution, we have estimated the corresponding distribution parameters consisting of shape and scale through data fitting. Clearly, these two parameter values are also pretty close for the generated data. All these results have demonstrated that the generated data for each wind farm shares very similar  characteristics as the actual data.

\vspace{3pt}
To sum up, this paper presented a wind power scenario generation approach based on GAN that can better incorporate the spatial correlation among multiple wind farms.
Our future research directions include using the generated scenarios in power system operations and planning tasks, as well as extending to develop scenario generation schemes for other energy resources.


\bibliographystyle{IEEEtran}
\bibliography{ref.bib}

\end{document}